%% file: main_neurips_format.tex
\documentclass{article}

\PassOptionsToPackage{numbers, compress}{natbib}
\usepackage[preprint]{neurips_2026}

\usepackage{times}
\usepackage{epsfig}
\usepackage{graphicx}
\usepackage{amsmath}
\usepackage{amssymb}
\usepackage[utf8]{inputenc} 
\usepackage[T1]{fontenc}    
\usepackage{url}            
\usepackage{booktabs}       
\usepackage{amsfonts}       
\usepackage{nicefrac}       
\usepackage{microtype}      
\usepackage{xcolor}
\usepackage{wrapfig}	
\usepackage[most]{tcolorbox} 
\usepackage{multicol} 
\usepackage[table]{xcolor}  
\usepackage{bbm} 
\usepackage[belowskip=0pt,aboveskip=4pt,font=small]{caption}
\usepackage[belowskip=0pt,aboveskip=0pt,font=small]{subcaption}
\usepackage{multirow}
\usepackage{tabularx}

\usepackage{enumerate}
\usepackage[olditem,oldenum]{paralist}

\usepackage{alltt}
\usepackage{listings}

\input{space_saver}
\usepackage{mysymbols}

\usepackage[pagebackref=true,breaklinks=true,letterpaper=true,colorlinks,bookmarks=false,citecolor=blue,linkcolor=green]{hyperref}

\newcommand{\dataset}{\mbox{TECCI}\xspace}
\newcommand{\overall}{\mbox{Overall}\xspace}
\newcommand{\verify}[1]{\textcolor{orange}{#1}}

\usepackage{xspace}
\newcommand{\nbp}{Nano Banana Pro\xspace}
\newcommand{\nbtwo}{Nano Banana 2\xspace}
\newcommand{\gpt}{GPT Image 1.5\xspace}
\newcommand{\grok}{Grok Imagine Pro\xspace}
\newcommand{\seedream}{Seedream 5.0 Lite\xspace}

\newcommand{\pz}{\hphantom{0}}

\begin{document}

\title{TECCI: Tricky Edits of Collected and Curated Images}


\author{Aishwarya Agrawal\thanks{Equal contribution, ordered alphabetically. $^\dagger$ Work partially done while Agrawal was at Google DeepMind.}$^{~,\dagger,1}$,
Roy Hirsch$^{*,1}$, 
Yasumasa Onoe$^{*,2}$, 
Sherry Ben$^2$, 
Jason Baldridge$^2$ \\
$^1$Google Research,
$^2$Google DeepMind\\
\url{https://google-deepmind.github.io/tecci}
}

%

\maketitle

\begin{abstract}
Despite tremendous recent progress, current text-guided image editing methods still struggle with many aspects of editing involving instruction following, minimally editing the source image, and ensuring high visual quality. These problems are especially apparent when the requested edit is challenging, such as those that involve position, motion, viewpoint, scale and creative edits. To systematically test generative image editors, we propose a novel image editing benchmark -- \dataset: \textbf{T}ricky \textbf{E}dits of \textbf{C}ollected and \textbf{C}urated \textbf{I}mages. \dataset consists of  a completely new set of images we are releasing.
The images in \dataset span 7 image categories. 
The images and these categories were curated intentionally to target weaknesses of existing methods. The edit instructions in \dataset are automatically generated by Gemini, covering 5 edit types per source image. We also curated a set of 530 images for which we created challenging manually written edit instructions. Overall, \dataset contains 7550 pairs of images and edit instructions.  

We conduct human evaluations of five leading image editing models on \dataset. Humans judge outputs along three dimensions: 1) instruction following, 2) minimality of the edits, and 3) visual quality. To scale-up the evaluation, we also build an auto-rater using Gemini 
that achieves $74.7\%$ accuracy in matching human evaluations. 
Our evaluations reveal that: 
1) none of the models exceed a 22\% overall success rate, demonstrating the challenging nature of \dataset, 
2) \nbp is the best performing model overall, 
3) models perform significantly better at instruction following compared to minimal edits and visual quality,  
4) models struggle with editing architecture and nature images which require strong understanding of spatial layout and intricate visual details. 
5) reasoning and creative edits are the most difficult, whereas color and appearance edits are the easiest.

\end{abstract}
\input{intro}
\input{related_work}
\input{dataset}
\input{benchmarking}

\input{conclusion}

{\small
\bibliographystyle{unsrtnat}
\bibliography{strings,references}
}
\clearpage
\input{appendix_overview}



\end{document}

%% file: space_saver.tex
\belowdisplayskip \abovedisplayskip

\newlength{\sectionReduceTop}
\newlength{\sectionReduceBot}
\newlength{\subsectionReduceTop}
\newlength{\subsectionReduceBot}
\newlength{\abstractReduceTop}
\newlength{\abstractReduceBot}
\newlength{\captionReduceTop}
\newlength{\captionReduceBot}
\newlength{\subsubsectionReduceTop}
\newlength{\subsubsectionReduceBot}

\newlength{\eqnReduceTop}
\newlength{\eqnReduceBot}

\newlength{\horSkip}
\newlength{\verSkip}

\newlength{\figureHeight}

%% file: intro.tex
\section{Introduction}
\label{sec:intro}

\begin{figure*}[t]
    \centering    \includegraphics[width=0.99\textwidth]{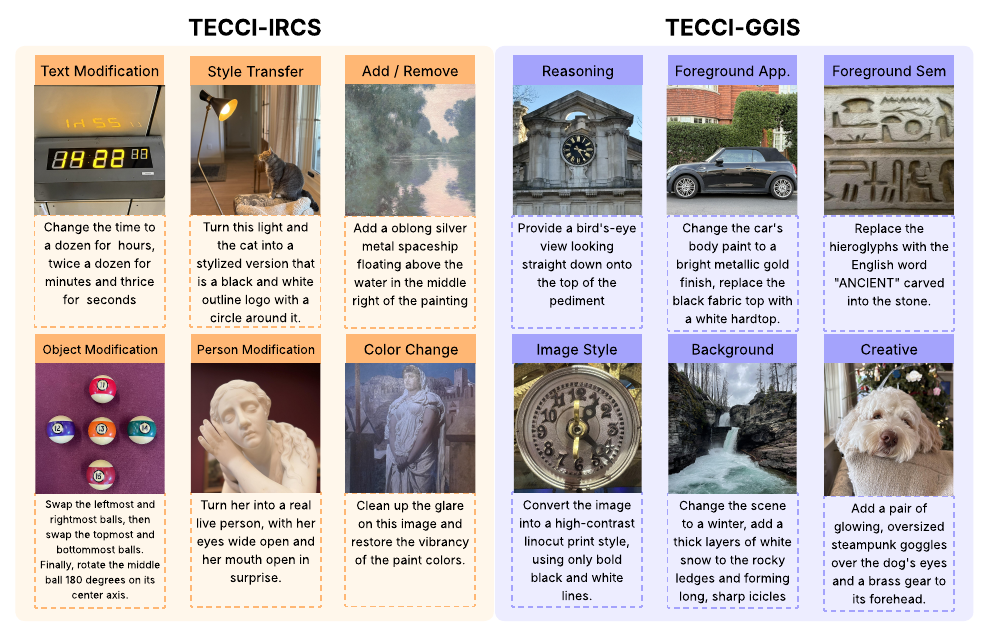}
    \caption{Overview of \dataset dataset. Representative samples from the different subsets and edit categories.}
    \label{fig:dataset_samples}
\end{figure*}

Text guided image editing is being increasingly used to benchmark the capabilities of generative image models \cite{magicbrush,aurora,earl,emuedit,omniedit,editbench,imagenworld}. This task evaluates whether these models can -- 1) \emph{faithfully} follow edit instructions (specified via a text prompt), 2) \emph{minimally} edit the source image, i.e., do not make any changes in the source image beyond what is specified in the instruction, 3) maintain the \emph{visual quality} of the source image in the generated edited image. Thus, this task serves as a good benchmark to evaluate visio-linguistic understanding as well as controlled generation capabilities of generative models.

We present a novel text-guided image editing benchmark to systematically evaluate generative models on challenging edit types: \textbf{T}ricky \textbf{E}dits of \textbf{C}ollected and \textbf{C}urated \textbf{I}mages (\dataset). \dataset targets image and edit types that have been observed (qualitatively) to be difficult for models, such as changing the text displayed on images, changing the time shown on clocks, structural and viewpoint changes to buildings, pose and action changes to animals and art subjects, state changes to vehicles, and scale changes to landscapes. \dataset also includes humorous creative edits that require models to carry out some degree of imagination. Unlike most existing image editing dataset \cite{magicbrush,aurora,earl,emuedit,omniedit,editbench,imagenworld}, the images \dataset have been taken and donated by the authors, ensuring that these images have not been seen by any current models in their training. 
\dataset spans 7 image categories: Text, Clock, Vehicle, Architecture, Art, Animal, and Nature. 
The edit instructions in \dataset are automatically generated by Gemini covering 5 edit types per source image. We also curated a smaller set (530) of challenging manually written edit instructions. Overall, \dataset consisting of 7550 pairs of images and edit instructions (see Fig. \ref{fig:dataset_samples} for representative examples).

We conduct human evaluations of five state-of-the-art image editing models on our proposed benchmark. Humans judge each edited sample along three dimensions: 1) faithfulness to the language instruction, 2) minimality of the edits, 3) visual quality of the edited image. In order to scale-up the evaluation, we also build an auto-rater using Gemini that achieves $74.7\%$ accuracy in matching human evaluations.
Our evaluations reveal that: 
1) none of the models exceed a 22\% overall success rate, demonstrating the challenging nature of \dataset, 
2) \nbp is the best performing model overall, 
3) models perform significantly better at instruction following compared to minimal edits and visual quality,  
4) models struggle with editing architecture and nature images which require strong understanding of spatial layout and intricate visual details. 
5) reasoning and creative edits are the most difficult, whereas color and appearance edits are the easiest.

\textbf{Contributions.} We make the following contributions:
\begin{compactitem}[\hspace{0pt}$\circ$]
\item We curate and release a large-scale challenging image editing benchmark -- \dataset -- consisting of 7550 pairs of images and edit instructions. The images in \dataset are completely new, manually captured and donated by the authors. 
\item We conduct a large-scale human evaluation of five state-of-the-art editing models on \dataset. In total, we collect 32,875 human judgements.
\item We conduct a comprehensive analysis of model performance using the collected human judgements revealing gaps in model capabilities and directions for future research.
\item We build an auto-rater to scale up the model evaluation on the full set of \dataset. We show that our auto-rater achieves an accuracy of $74.7\%$ in matching human evaluations.

\end{compactitem}

%% file: related_work.tex
\section{Related Work}
\label{sec:related_work}

Several benchmarks have been proposed for the task of text guided image editing \cite{magicbrush,aurora,earl,emuedit,omniedit,editbench,imagenworld}. These benchmarks differ along the following axes: 
\begin{compactitem}[\hspace{0pt}$\bullet$]
\item \textbf{Localized edits vs global edits}: Localized edits involve edits to a local region of an image, spanning simple edits such as object addition and removal, attribute changes to an object, and complex edits such as counting, spatial relations and action edits. Global edits involve making global changes to an image such as stylistic changes, changing the weather, daytime into nighttime etc. \dataset covers all these types of edits.
\item \textbf{Image source}: The images in the existing benchmarks are sourced either from existing natural image datasets such as COCO \cite{Chen2015MicrosoftCC}, OpenImages \cite{OpenImages}, Visual Genome \cite{krishna2017visual} or are generated synthetically using image generation models. \dataset differs from all existing benchmarks in that \dataset uses completely new set of images manually taken by one of the authors. Thus, \dataset includes natural images that do not already exist on the internet.
\item \textbf{Edit instruction source}: The edit instructions in the existing benchmarks are either manually written or automatically generated using LLMs. \dataset includes a larger set of automatically generated instructions as well as a smaller challenging set of instructions manually written by one of the authors. 
\item \textbf{Dataset size}: The sizes of the existing datasets range from 250 to 10,000. \dataset lies on the higher end of this spectrum, with 7.5K samples.
\item \textbf{Availability of human judgements}: Some existing benchmarks collect human judgements on the model generated edits, either using single-sided (judging each model output independently) or side-by-side (comparing outputs from two different models) evaluations. \dataset conducts a large scale single-sided human evaluation on 1,315 samples on 5 model outputs using 5 raters, resulting in a total of 32,875 collected human judgements. This is significantly larger than the scale of human studies in most existing benchmarks.
\end{compactitem}

%% file: dataset.tex
\section{\dataset: Dataset}
\label{sec:dataset}

\subsection{Image Collection and  Curation}
\label{sec:image_collection_and_curtion}

\dataset consists of newly collected images that have never appeared on the internet. To collect them, one of the authors took photographs over a multiyear period, with the aim of creating a new corpus of natural images for testing next generation image generation and editing models. The selection process involved a mix of identifying scenes that captured known failure modes of models (e.g. complex text rendering), opportunistic identification of interesting scenes and subjects and generally building a body of images that would provide the raw material for benchmarking. These include:

\begin{compactitem}[\hspace{0pt}$\bullet$]
    \item \textbf{Text}: This category evaluates a model's ability to render accurate typography within a scene. Beyond mere legibility, it tests for spelling accuracy, font consistency, and how naturally the text maps onto varied surfaces (such as curved labels, book covers, or weathered street signs).
    \item \textbf{Clock}: Clocks serve as a strict benchmark for spatial reasoning and logical geometric alignment. They have precise circular or digital layouts. Successful depiction relies on the hands being positioned logically (e.g., the hour hand properly reflecting the minute hand's progress).
    \item \textbf{Vehicle}: Vehicles are also inherently interesting for multiple reasons, including the specificity of type (van vs car), make (Ford vs BMW), model (Mustang vs Focus), trim (911 GTS vs 911 RS) and the many relevant properties (color, size, tire types, words, designs) and the importance of specific view points (front three quarters, rear overhead) when describing and depicting them.
    \item \textbf{Architecture}: Buildings are interesting in a similar respect, but they are more varied than vehicles (less standardized). They also appear regularly in real life and in classic paintings.
    \item \textbf{Art}: This category benchmarks compositional synthesis and stylistic replication across diverse media (e.g., oil, watercolor). While the TECCI dataset excludes photographs of people, the art category leverages the historical depth of paintings and statues to evaluate models' ability to accurately understand and depict the human form, including complex poses.
    \item \textbf{Animal}: Animals provided another rich source of real life settings due to their poses and settings, plus the world knowledge needed to identify and depict different species and breeds.
    \item \textbf{Nature}: To capture organic real-world environments, nature scenes test a model's ability to render complex textures, lighting, and diverse landscapes without the predictable, rigid geometry of man-made objects.
\end{compactitem}

Further details on image curation (licensing, post processing etc.) are provided in Appendix \ref{app:image-curation}.

\subsection{Edit Instruction Curation}
\label{sec: edit_instruction_curation}

To curate edit instructions, we divided the images into two sets: TECCI-IRCS and TECCI-GGIS. Below we provide further details about each of these sets.

\textbf{TECCI-IRCS}, short for TECCI - Instruction Rich Challenge Set. 
This set consists of challenging manually written edit instructions. 
The images in this set were manually selected, identifying those which provided opportunities for creating edit requests expected to be hard. 
Then, instructions were manually written by the authors as challenging edits. These instructions span the following categories:

\begin{compactitem}[\hspace{0pt}$\bullet$]
    \item \textbf{Text Modification}: High-precision edits involving adding, altering, or translating written text and typography, including specific font styling and numerical data.
    \item \textbf{Style Transfer}: Shifting the artistic medium or aesthetic of an image, such as transforming a conceptual sketch into a photorealistic output or applying specific painterly styles.
    \item \textbf{Object Addition Removal}: The seamless insertion of new subjects (people, animals, or objects) or the erasure of existing elements through high-fidelity background inpainting.
    \item \textbf{Object Modification}: Physical manipulation of existing items, including resizing, rotating, swapping, or fundamentally altering their material properties and state.
    \item \textbf{Person Modification}: Fine-grained adjustments to human subjects, ranging from physical appearance (age, clothing, features) to specific postures, gestures, and gaze directions.
    \item \textbf{Subject Recontextualization}: Transplanting a subject from its original environment into an entirely new setting or an unexpected, complex scenario.
    \item \textbf{Color Change}: Targeted color swaps for specific architectural or object-level elements, as well as global modifications to the scene's lighting conditions and mood.
\end{compactitem}

\textbf{TECCI-GGIS}, short for TECCI - Gemini Generated Instruction Set. This set includes all the rest of the images, that were not included in TECCI-IRCS. The edit instructions for the images in this set were generated automatically using Gemini 3 Pro \cite{nanobananapro}.
We prompted Gemini to generate edit instructions for specific types of edits depending on the image category, such as, text translation edits for text images, time edits for clock images, state and motion edits for vehicle images, viewpoint edits for architecture images, pose and action edits for human and animal images, scale edits for nature images etc. These edit types were designed manually by one of the authors based on their prior experience of challenging edits for SOTA models. We also include a creative edit type for each image which requires a certain degree of imagination and is humorous in nature. Overall, we designed 5 edit types for each image category. Please see the Appendix (Table \ref{tab:ggis_edits}) for the full list of these edit types per image category. We further categorized these image-category-specific-edit types into the following meta-edit types. Each of the meta-edit types spans multiple image categories.

\begin{compactitem}[\hspace{0pt}$\bullet$]
    \item \textbf{Reasoning}: Edits requiring complex reasoning such as text translation in text images, text modification in clocks and vehicles, viewpoint changes in architectures, art restoration, position changes in animals, scale changes in nature (e.g., zooming into a specific mountain).
    
    \item \textbf{Foreground Appearance}: Edits targeting appearance changes to the foreground object such as style changes to the text in text images, color/texture/material changes to clocks, vehicles, and architectures, adding color to monochrome art images, and color/texture changes to animals.  
    
    \item \textbf{Foreground Semantics}: Edits targeting semantic changes to the foreground object such as text content changes in text images, time changes in clocks, state and motion changes in vehicles, structural changes in architectures, pose/action/expression changes in art subjects, object addition/removal in nature images, and pose and action changes in animals. 
    
    \item \textbf{Image Style}: Edits targeting global stylistic changes to the images of clocks, arts, animals and nature (e.g., converting to: comic style, photorealistic image, specific kind of painting etc.)
    
    \item \textbf{Background}: Edits targeting background changes, e.g. a new background in text images while preserving text content and style, environment/theme changes to vehicles and buildings while preserving the vehicle/building, and changes in landscapes (e.g., changing the weather).
    
    \item \textbf{Creative}: Creative and humorous changes to the main subject of the image -- text/clock/vehicle/architecture/art subject/animal/nature elements. 
\end{compactitem}

\subsection{Image Caption Curation}

We generate a detailed caption for each image using Gemini 2.5 Flash \citep{comanici2025gemini25pushingfrontier}, prompting the model to describe salient visual features including key objects, their spatial relationships, and lighting conditions. This process yields one caption per image across the full dataset of 1,934 images. In this work, we use these captions as textual features for the image-type clustering described in Section~\ref{sec:image_collection_and_curtion}.
In addition to these detailed descriptions, we also generate a short caption for each image using the same model, prompting it to produce a concise, high-level summary of the image content.
These captions provide textual metadata useful for future research on the evaluation of text-to-image generation, image captioning and image-text alignment.

\begin{wraptable}{r}{6.3cm}
\centering
\vspace{-13pt}
\caption{Summary statistics of the TECCI dataset.}
\label{tab:summary_stats}
\footnotesize
\begin{tabular}{lrrr}
\toprule
\textbf{Statistic} & \textbf{IRCS} & \textbf{GGIS} & \textbf{Total} \\
\midrule
Images & 530 & 1,404 & 1,934 \\
Edit instructions & 530 & 7,020 & 7,550 \\
Instructions / image & 1 & 5 & 3.9 \\
Unique edit types & 7 & 21 & 28 \\
Unique image types & 7 & 7 & 7 \\
\midrule
\multicolumn{4}{l}{\textit{Instruction Length (words)}} \\
\quad Min & 4 & 8 & 4 \\
\quad Max & 102 & 61 & 102 \\
\quad Mean & 29.0 & 22.8 & 23.2 \\
\quad Std & 15.5 & 5.2 & 6.7 \\
\bottomrule
\end{tabular}
\vspace{-10pt}
\end{wraptable}

\subsection{Dataset Analysis}
\label{sec: dataset_analysis}

\textbf{Dataset Overview}
We provide a comprehensive analysis of the TECCI dataset to characterize its scale, diversity, and linguistic properties.
Table~\ref{tab:summary_stats} summarizes the statistics of our dataset.
TECCI comprises two complementary subsets: TECCI-GGIS, which contains 1,404 images paired with 7,020 edit instructions (five instructions per image), and TECCI-IRCS, which contains 530 images with 530 edit instructions (one per image). 
TECCI spans 7 image categories and 28 edit types (7 IRCS and 21 GGIS). Instructions average 23.2 words, with manual IRCS prompts being longer (29.0 words) on average than the LLM generated GGIS ones (22.8 words).
In total, TECCI provides 1,934 unique images and 7,550 edit instructions.

\begin{figure}[t]
\centering
\includegraphics[width=1\linewidth]{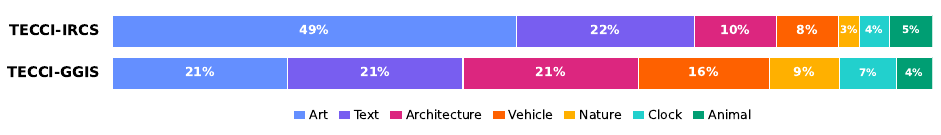}
\caption{Image type distributions. TECCI-IRCS prioritizes specific challenging edits (resulting in an unbalanced distribution). TECCI-GGIS utilizes uniform sampling to achieve a balanced distribution across image types.}
\vspace{-10pt}
\label{fig:image_type_dist}
\setlength{\belowcaptionskip}{-10pt}
\end{figure}

\textbf{Image Type Distribution}
Figure~\ref{fig:image_type_dist} shows the composition of the image types described in Section~\ref{sec:image_collection_and_curtion} for TECCI-IRCS and TECCI-GGIS. For TECCI-IRCS, images are selected purely based on their potential for challenging edits, such as rare combinations of objects and their attributes (e.g., materials, colors, and shapes), resulting in more dominant represention of certain categories such as Art . In contrast, for TECCI-GGIS, images are sampled more uniformly across different types, resulting in a more balanced distribution.

\textbf{Edit Instructions Distribution}
Figure~\ref{fig:edit_type_dist} depicts the distribution of edit instructions across the IRCS categories and GGIS meta-categories described in Section~\ref{sec: edit_instruction_curation}. Importantly, the edit type – image type co-occurrence heatmap (Figure~\ref{fig:image_type_heatmap}) reveals that edit types cover several image categories rather than being concentrated on one or two. For example, Reasoning, Foreground Semantics and Creative edits in GGIS cover all 7 image categories. Similarly, This cross-category coverage challenges models’ generalization capabilities across both editing operations and visual contexts simultaneously.

Linguistic complexity analysis shows that TECCI-IRCS exhibits greater lexical diversity per instruction: while TECCI-GGIS has a larger total vocabulary (6,115 unique tokens) compared to TECCI-IRCS (2,253 unique tokens), TECCI-IRCS exhibits a higher Type-Token Ratio (the ratio of unique words to total words; 0.1461 vs. 0.0377) and a larger proportion of Hapax Legomena (words appearing only once; 52.9\% vs. 37.1\%).

%% file: benchmarking.tex
\section{Benchmarking Image Editing Models on \dataset}
\label{sec:benchmarking}

\begin{figure}[t]
\centering
\includegraphics[width=1\linewidth]{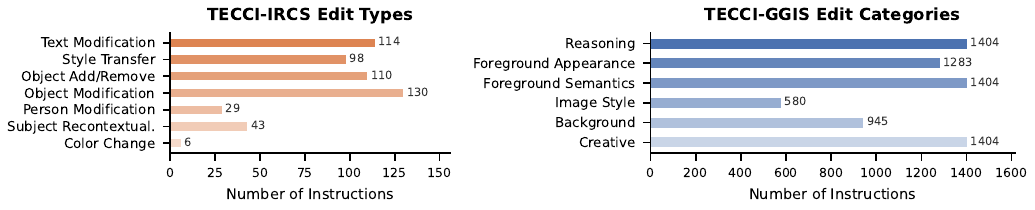}
\caption{Distribution of edit categories across TECCI-IRCS (left) and TECCI-GGIS (right).}
\vspace{-10pt}
\label{fig:edit_type_dist}
\setlength{\belowcaptionskip}{-10pt}
\end{figure}

\begin{figure}[t]
\centering
\includegraphics[width=1\linewidth]{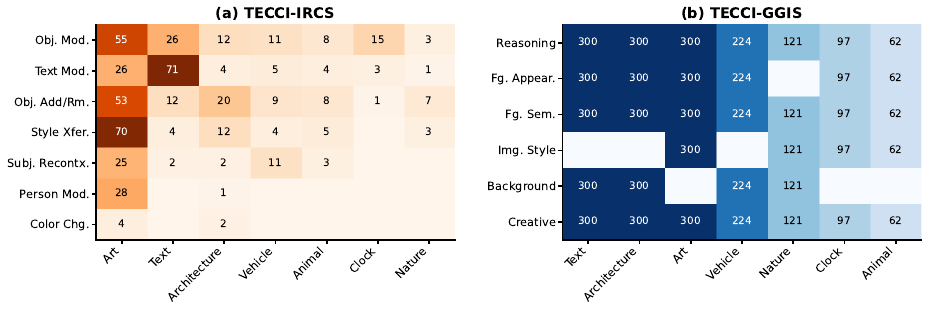}
\caption{Co-occurrence heatmaps of edit categories and image types in (a) TECCI-GGIS and (b) TECCI-IRCS. Cell values indicate instruction counts; color intensity reflects frequency.}
\label{fig:image_type_heatmap}
\setlength{\belowcaptionskip}{-10pt}
\end{figure}

\subsection{Models Evaluated}
\label{sec:model_evaluation}
We evaluate the performance of several state-of-the-art image generation models using the \dataset dataset.\footnote{State-of-the-art models were selected based on their rankings on the Arena AI Image Editing Leaderboard as of May 1, 2026. Available at: https://arena.ai/leaderboard/image-edit.} We report the performance of the following proprietary models: \textit{\nbtwo} \citep{nanobananaflash}, \textit{\nbp} \citep{nanobananapro}\footnote{\nbtwo and \nbp are referred to the Gemini 3.1 Flash Image Preview and Gemini 3 Pro Image Preview models, respectively, as accessed via the official Gemini API.}, \textit{\grok} \citep{grok}, \textit{\seedream} \citep{seedream} and \textit{\gpt} \citep{gptimage}. The outputs for all proprietary models were obtained directly through their respective official APIs with default hyper-parameters for maximal generation quality. Representative samples are presented in Figure \ref{fig:cat}, while additional samples are provided in Appendix \ref{app:additional_samples}.

\subsection{Evaluation Criteria}
\label{sec: eval_criteria}

Evaluating the quality of instruction-based image editing is challenging \citep{ku2023viescore, sani2026imagenworld}. It is inherently multi-dimensional and subjective, so we defined granular scoring guidelines. 
To isolate and capture the various facets of image editing performance, our pipeline utilizes three distinct evaluation criteria: \textbf{1. Instruction Following (IF):} Assesses the semantic alignment between the edit instruction and the resulting edited image. It measures whether the generative model accurately and completely fulfilled all the required modifications, serving as the primary benchmark for the system’s functional utility. \textbf{2. Image Consistency (IC):} Evaluates the preservation of the original image’s identity and non-targeted regions and elements. This criterion penalizes ``over-editing'' or unrequested alterations to the background and perspective, ensuring the edit is minimal. \textbf{3. Visual Quality (VQ):} Captures the aesthetic and technical excellence of the edit, identifying any introduced artifacts such as blurring, pixelation, or unnatural blending. It determines if the modifications are seamlessly integrated to maintain a realistic, high-resolution appearance.

\subsection{Human Evaluation on a subset of TECCI}
\label{sec: human_evals}

\textbf{Curating Human Judgements}: We conducted a comprehensive study to collect human judgement scores for all 5 models mentioned in Sec \ref{sec:model_evaluation}. A dataset of 1,315 pairs of images and edit instructions was drawn from the TECCI benchmark: 1) 1050 from GGIS: drawing 30 images uniformly at random from each of the 7 image categories, and including all 5 edit instructions per image, 2) 265 pairs of images and edit instructions from IRCS drawn uniformly at random.
Each generated image was subjected to review by five independent raters, resulting in a total of 32,875 granular assessments (across all 5 models). Raters were required to provide independent assessments for each of the three evaluation criteria (Sec \ref{sec: eval_criteria}) to ensure a multi-dimensional perspective on model performance. Each criterion was quantified using a five-point Likert scale, ranging from \textit{`Not at all' (1)} to \textit{`Completely' (5)}, allowing for a granular measurement of the model's successes and failures. 
To ensure inter-rater reliability and alignment, all participants underwent comprehensive training and followed a detailed set of scoring guidelines. The evaluation template is provided in Appendix~\ref{app:human-eval-sinlge-temp}. 
To ensure data quality, we further filter raters by removing those with a low Pearson correlation ($r < 0.1$), which excluded 10–17\% of the total ratings. Inter-rater reliability was measured using \textit{Krippendorff’s $\alpha$} (interval), yielding a mean coefficient of 0.58 (averaged across the different models and evaluation criteria; for per-criterion Krippendorff’s $\alpha$ see Appendix Tab \ref{tab:ar_corr}). This indicates a moderate degree of agreement, which is consistent with the inherent subjectivity of image editing evaluation. 

\textbf{Success Rate Metric}: To evaluate the models via human judgment, we define a success rate metric, which is computed both at the per-criterion and aggregate levels. The per-criterion success rate represents the proportion of samples whose average rating for that criterion, averaged across the five human evaluators, exceeds a predefined threshold $\tau$. The aggregate success rate metric \emph{\overall (O)} is defined as: $O = \mathbbm{1} (IF \ge \tau \land IC \ge \tau \land VQ \ge \tau)$ where $\mathbbm{1}$ is the indicator function, $IF$, $IC$, $VQ$ represent the per-criterion average rating (across the five human evaluators), and $\tau$ is a predefined threshold. This, for an edit to be considered successful, it should score at least $\tau$ on each of the three criteria.
Empirically, we found that setting $\tau{=}4.5$ ensures high quality fulfillment of the three evaluation criteria (success rates based on other thresholds are presented in Appendix \ref{app:human-evaluation}). Our analysis centers on this metric, as it provides a clear benchmark for the perfect editing rate \footnote{Statistics on raw human scores are presented in Appendix \ref{app:human-evaluation}}.

\begin{table*}[ht!]
\centering
\caption{Human-based success rate (\%) on \dataset benchmark, across IRCS and GGIS subsets. IF = Instruction Following, IC = Image Consistency, VQ = Visual Quality, O = \overall.}
\label{tab:human_baseline_results}
\small
\begin{tabular}{@{}l cccc cccc cccc@{}}
\toprule
\multirow{2}{*}{\textbf{Model \textbackslash{} Dataset}} & \multicolumn{4}{c}{\textbf{IRCS}} & \multicolumn{4}{c}{\textbf{GGIS}} & \multicolumn{4}{c}{\textbf{IRCS + GGIS}} \\
\cmidrule(lr){2-5} \cmidrule(lr){6-9} \cmidrule(l){10-13}
& \textbf{IF} & \textbf{IC} & \textbf{VQ} & \textbf{O} & \textbf{IF} & \textbf{IC} & \textbf{VQ} & \textbf{O} & \textbf{IF} & \textbf{IC} & \textbf{VQ} & \textbf{O} \\ 
\midrule
Nano Banana Pro          & \cellcolor{gray!20}39.0 & \cellcolor{gray!20}34.5 & 36.0 & \cellcolor{gray!20}19.7 & 47.7 & \cellcolor{gray!20}44.0 & \cellcolor{gray!20}38.5 & \cellcolor{gray!20}23.0 & 46.0 & \cellcolor{gray!20}42.1 & \cellcolor{gray!20}38.0 & \cellcolor{gray!20}22.3 \\
\nbtwo      & 37.7 & 30.6 & \cellcolor{gray!20}37.7 & 17.7 & 47.6 & 36.5 & 33.6 & 20.2 & 45.6 & 35.3 & 34.4 & 19.7 \\
\gpt         & 29.9 & \pz8.3  & 20.8 & \pz4.9 & 45.0 & \pz9.9  & 21.1 & \pz5.9  & 42.0 & \pz9.6  & 21.0 & \pz5.7  \\
Grok Imagine Pro      & 37.4 & 33.2 & 33.6 & 14.7 & \cellcolor{gray!20}49.3 & 40.8 & 37.3 & 21.3 & \cellcolor{gray!20}46.9 & 39.2 & 36.6 & 20.0 \\
\seedream     & 32.6 & 29.5 & 21.1 & 10.3 & 43.6 & 40.6 & 30.3 & 18.2 & 41.4 & 38.4 & 28.5 & 16.7 \\

\bottomrule
\addlinespace[1ex]
\end{tabular}
\end{table*}

\textbf{Overall Performance}: Table~\ref{tab:human_baseline_results} summarizes the primary results of the human evaluation study. We report the \overall success rate alongside the per-criterion success rates. The low \overall scores across the model suite highlight the inherent difficulty of \dataset. Even the most capable models struggle to exceed a 22.3 overall success rate. IRCS stands out as a significantly more challenging subset. The top-performing model achieves only a 19.7 \overall score, in comparison to 23.0 achieved on the GGIS subset. This gap stems from the increased difficulty of the manually curated instruction set, which requires greater creative variance and complexity. \nbp demonstrates superior overall performance, followed by Grok and \nbtwo. Notably, \nbp leads across most individual dimensions, with the exception of instruction following, for which Grok obtains the highest score. While \gpt achieves competitive IF scores, it significantly underperforms compared to other models across all remaining evaluation dimensions. This discrepancy is largely attributable to a tendency toward over-editing and lower overall output quality. 

\begin{wraptable}{r}{5cm}
\centering
\vspace{-13pt}
\caption{Comparative Elo Ratings (99\% CI) for IRCS and GGIS Datasets.}
\label{tab:elo_ratings}
\footnotesize
\setlength{\tabcolsep}{3pt}
\begin{tabular}{lcc}
\toprule
\textbf{Model} & \textbf{IRCS Elo} & \textbf{GGIS Elo} \\
\midrule
Nano Banana 2 & $1055 \pm 7$ & $1036 \pm 6$ \\
Nano Banana Pro & $1045 \pm 7$ & $1031 \pm 7$ \\
Grok Imagine Pro & \pz$987 \pm 7$ & \pz$998 \pm 7$ \\
GPT Image 1.5 & \pz$971 \pm 7$ & \pz$969 \pm 7$ \\
Seedream 5.0 Lite & \pz$941 \pm 7$ & \pz$967 \pm 7$ \\
\bottomrule
\end{tabular}
\vspace{-10pt}
\end{wraptable}

\textbf{Side by Side Model Comparison}: We further conducted side by side comparisons between model pairs, in which we asked humans to indicate which model's output they prefer (see Appendix \ref{app:human-eval-sxs-temp} for the detailed template). In this study, every pair of images from two tested models is judged by three distinct raters - yield an extensive comparative analysis. Table \ref{tab:elo_ratings} gives the Elo score \citep{zheng2023judging} (with 99\% CI)  for each model as the main metric of this study.
In this evaluation, aspire to capture humans overall preference between two models. Hence we did not define the specific criteria that we defined in the single sided evaluations (Sec. \ref{sec: eval_criteria}). 
Ultimately, side-by-side evaluations offer a vital relative comparison that complements single-sided assessments. This analysis reveals similar trends as \nbtwo and \nbp achieve highest performance across both sets of \dataset, significantly outperforming the competitors. However, unlike the single sided evaluation, Seedream 5.0 Lite receives the lowest scores in this setup, falling behind \gpt, particularly on the IRCS dataset where they scored 941 and 971, respectively. This divergence could stem from the differences between: 1) capturing overall vs. criteria specific judgements in side-by-side vs single-sided evaluations, 2) judging relative vs absolute model competence in side-by-side vs single-sided evaluations.

\textbf{Performance Comparison across the Evaluation Criteria}: While specific dimensions like Instruction Following (IF) reach the 40–50 range, Image Consistency (IC) and Visual Quality (VQ) remain lower. Furthermore, the aggregate \overall scores consistently trail individual dimension, indicating that while models may excel in a specific criteria, only a small fraction of generations achieve perfectness across all evaluation axes. The reported performances highlight the fact that while current models possess a robust capacity for instruction following, they struggle substantially with minimally editing and maintaining the visual quality of the source image. \nbp demonstrates the most balanced multidimensional profile, maintaining high IC scores (e.g., 42.1 in the combined dataset) without sacrificing VQ or IF, leading to the highest \overall of 22.3. In contrast, while Grok Imagine Pro achieves the top IF score of 46.9 on the combined dataset, its slightly lower VQ (36.6) and IC (39.2) prevent it from overtaking the lead in overall performance. \gpt IC underperforms with a score of 9.6—nearly five times lower than its instruction-following capability. This pronounced divergence highlights a critical bottleneck in maintaining minimal edits and visual fidelity relative to a reference image.

\begin{figure}[ht!]
     \centering
     \includegraphics[width=1\textwidth]{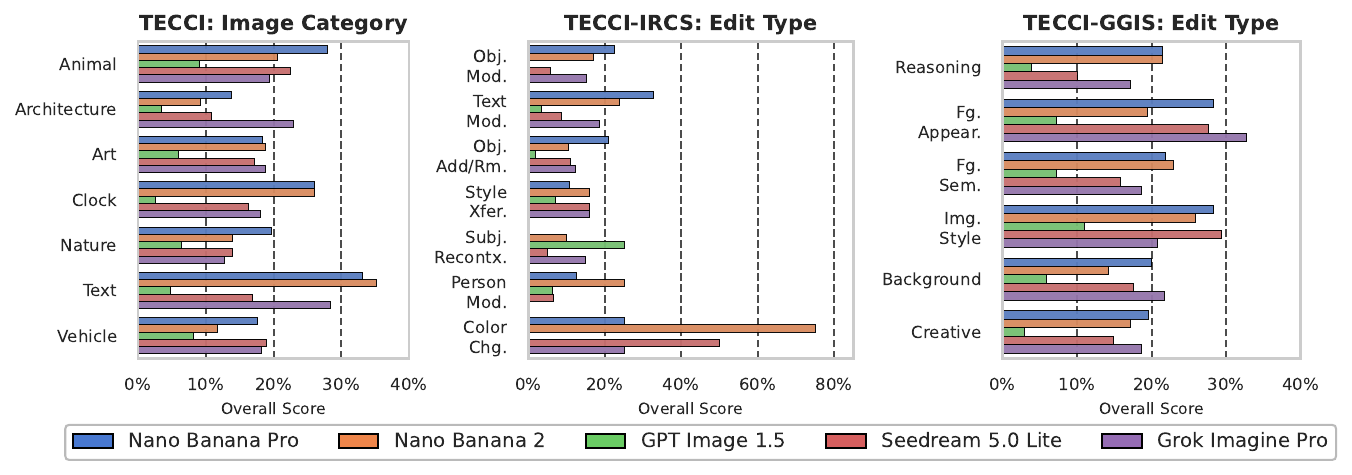}
     \caption{Comparative analysis of model performance across image categories and edit types. Reporting human-based \overall score for all the tested models across source image categories and edit types.}
     \label{fig:scores_dists}
\end{figure}

\textbf{Performances Comparison across the Image and Edit Types}: Figure \ref{fig:scores_dists} illustrates performance  by source image categories and edit instruction types (several models yielded an \overall score of zero for certain IRCS edit types). \textit{Text} and \textit{Animal} source image categories produce higher scores, likely because those are object-centric images that often feature isolated subjects that simplify the generative task. In contrast, \textit{Architecture} and \textit{Nature} present a substantially greater challenge, with even the best models rarely surpassing a 20\% success threshold. These categories typically demand superior spatial precision and a strong understanding of intricate visual details. For the IRCS subset, \textit{Color Change} is the easiest with \nbtwo reaching nearly 80\%, whereas \textit{Style Transfer}, \textit{Object Addition Removal}, \textit{Object Modification}, \textit{Person Modification}, and \textit{Subject Recontextualization} remain universally difficult, generally falling below the 20\% mark. 
For the GGIS subset, while \textit{Foreground Appearance} and \textit{Image Style} peak above 30\%, \textit{Reasoning}, \textit{Foreground Semantics}, \textit{Background} and \textit{Creative} plateau at approximately 20\%, highlighting a persistent struggle on complex reasoning and imagination. Overall, current state-of-the-art models demonstrate greater proficiency in executing low-level modifications, such as color and appearance changes, than on high-level edits that require complex reasoning or imagination. 
Lastly, the Nano Banana series and Grok Imagine Pro consistently outperform other models across the image and edit categories, whereas \gpt consistently under-performs with scores often trailing below 10\%.

\noindent 

\begin{figure*}[h]
    \centering
    \includegraphics[width=0.99\textwidth]{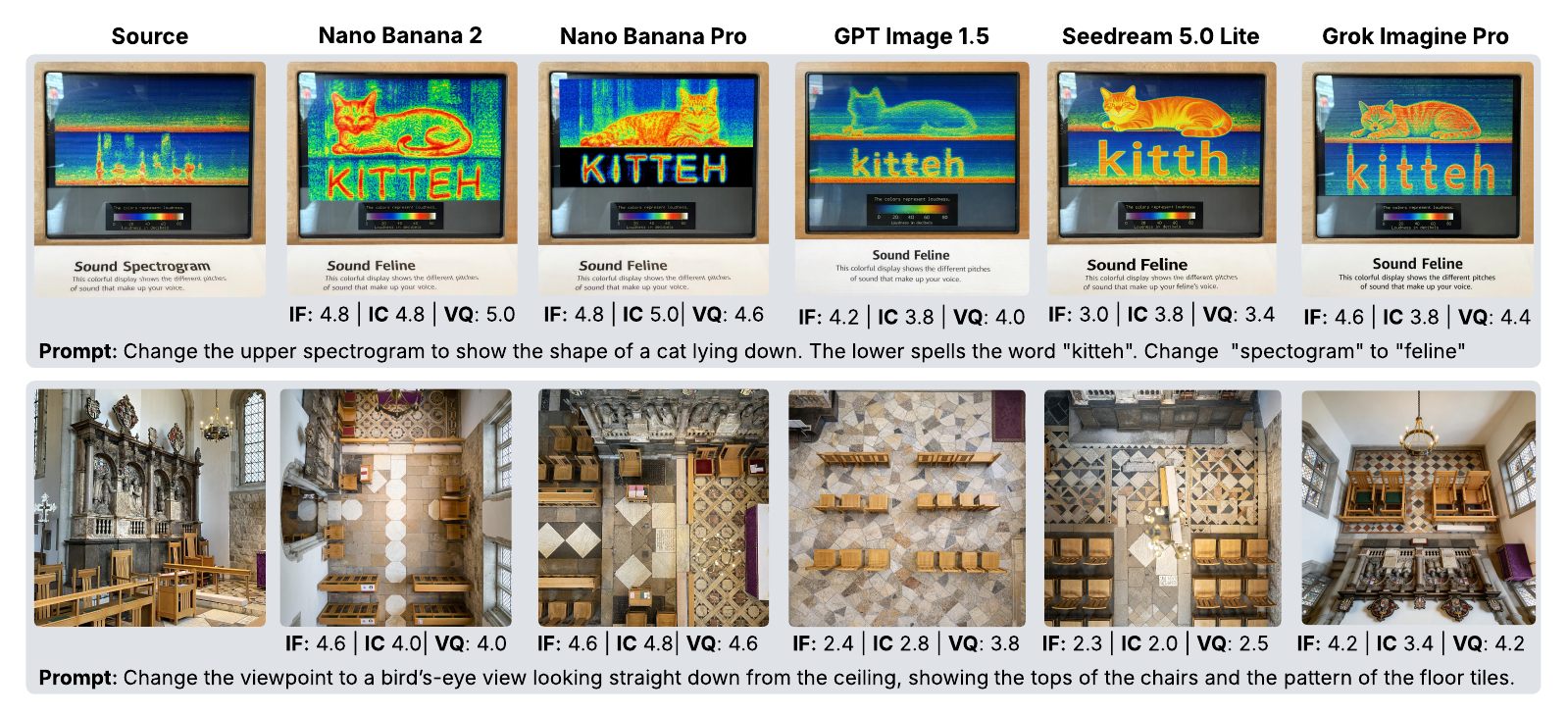}
    \caption{Sample images from TECCI-IRCS (top) and TECCI-GGIS (bottom). Each row includes the source image, the editing instruction, and the resulting edited images with their respective mean human scores.}
    \label{fig:cat}
\end{figure*}

\subsection{Automatic Evaluation on the Full set of \dataset}
\label{sec: autorater}

\begin{table*}[ht!]
\centering
\caption{Autorater-based success rate (\%) on all \dataset benchmark, across IRCS and GGIS subsets. IF = Instruction Following, IC = Image Consistency, VQ = Visual Quality, O = \overall.}
\label{tab:auto_baseline_results}
\small
\begin{tabular}{@{}l cccc cccc cccc@{}}
\toprule
\multirow{2}{*}{\textbf{Model \textbackslash{} Dataset}} & \multicolumn{4}{c}{\textbf{IRCS}} & \multicolumn{4}{c}{\textbf{GGIS}} & \multicolumn{4}{c}{\textbf{IRCS + GGIS}} \\
\cmidrule(lr){2-5} \cmidrule(lr){6-9} \cmidrule(l){10-13}
& \textbf{IF} & \textbf{IC} & \textbf{VQ} & \textbf{O} & \textbf{IF} & \textbf{IC} & \textbf{VQ} & \textbf{O} & \textbf{IF} & \textbf{IC} & \textbf{VQ} & \textbf{O} \\ 
\midrule
Nano Banana Pro   & 53.8 & \cellcolor{gray!20}36.0 & 46.2 & \cellcolor{gray!20}18.6 & 61.0 & \cellcolor{gray!20}45.0 & 38.3 & 20.1 & 59.5 & \cellcolor{gray!20}43.2 & 39.9 & \cellcolor{gray!20}19.8 \\
Nano Banana 2     & \cellcolor{gray!20}64.2 & 29.4 & \cellcolor{gray!20}49.4 & 16.2 & \cellcolor{gray!20}66.3 & 39.0 & 43.9 & \cellcolor{gray!20}20.4 & \cellcolor{gray!20}65.9 & 37.1 & \cellcolor{gray!20}45.0 & 19.5 \\
GPT Image 1.5     & 44.7 & \pz5.3  & 40.9 & \pz4.5  & 58.0 & \pz5.4  & 27.2 & \pz4.8  & 55.4 & \pz5.4  & 30.0 & \pz4.7  \\
Grok Imagine Pro  & 55.8 & 31.7 & 45.7 & 18.5 & 61.1 & 39.8 & \cellcolor{gray!20}44.1 & 19.5 & 60.1 & 38.2 & 44.4 & 19.3 \\
Seedream 5.0 Lite & 45.2 & 34.9 & 41.4 & 15.7 & 56.6 & 44.7 & 38.9 & 19.4 & 54.3 & 42.7 & 39.4 & 18.6 \\
\bottomrule
\addlinespace[1ex]
\end{tabular}
\end{table*}

Although human evaluation serves as the gold standard for assessing visual quality, it is bottlenecked by scalability constraints and inherent rater subjectivity. To overcome these limitations, we propose an MLLM-based automatic evaluation framework that enables streamlined and reproducible benchmarking on the \dataset dataset. The autorater processes the source image, the edit instruction, and the resulting edited image. We prompt the model to perform a rigorous, systematic analysis of the visual output, independently rating each of the three evaluation criteria on a 1–5 Likert scale. To ensure consistency and alignment, the model is provided with comprehensive guidelines regarding the experimental setup and a granular rubric for each scoring criterion. The complete prompt is detailed in Appendix \ref{app:auto-eval}. We employ Gemini 3 Flash \citep{gemini3flash} as our evaluation backbone. To mitigate stochastic variance, we perform five independent inference runs per sample and compute the mean.
Similar to human evaluation, we compute a model's success rate using the \overall metric on the mean autorater scores.
Our framework obtains strong alignment with human judgment, achieving $74.7\%$ accuracy in matching human-derived overall score. 

To calibrate the autorater's score distribution against that of the human raters, we employ Z-score normalization (mean-variance matching) relative to the human distribution, independentally for each criterion (see Appendix \ref{app:auto-eval} for details). Following this normalization, we apply $\tau{=}4.5$ to calculate success rates for the individual criteria and the \overall score. Table \ref{tab:auto_baseline_results} presents the success rates across the entire \dataset benchmark as measured by the autorater. The observed trends largely align with the human evaluation. Notably, \overall scores do not exceed 19.8, and both \nbp and \nbtwo consistently outperform other models across most setups. However, specific discrepancies exist. The autorater is biased to predict higher scores for IF and VQ compared to the human judges. Nevertheless, the \overall metric is aligned with human-based analysis and remains a robust indicator for ranking models when human ratings cannot be obtained.

%% file: conclusion.tex
\section{Conclusion}
\label{sec:conclusion}
We curate a novel and challenging image editing benchmark dataset, \dataset, consisting of 7550 pairs of images and edit instructions. \dataset consists of a completely new set of 1,934 images, unseen by current generative models. The images and edit instructions in \dataset are carefully curated by the authors to target the weaknesses of the SOTA image editing models. We conduct a large scale human evaluation of 5 SOTA image editing models on a subset of \dataset and further scale up the evaluation on the full dataset using a custom built auto-rater. Our evaluations reveal interesting gaps in model capabilities and directions for future research.
We hope that \dataset images and associated meta-data such as the automatically generated captions will serve as a useful resource for future research on the evaluation of text-to-image generation, image captioning and image-text alignment.

\textbf{Acknowledgements.} We would like to thank Radu Soricut and Jordi Pont-Tuset for their constructive feedback and support of our work.

%% file: appendix_overview.tex
\section*{Appendix}
\label{sec:appendix_overview}

\noindent The appendix consists of the following sections:
\begin{compactitem}[\hspace{0pt}$\bullet$]
\item Limitations and Future Work (Sec \ref{app:limitations})
\item Broader Impact (Sec~\ref{app:broader_impact})
\item Computational Requirements (Sec~\ref{app:resources})
\item Additional Details on Image Collection and Curation (Sec~\ref{app:image-curation})
\item Full List of TECCI-GGIS Edit Types (Sec \ref{app:ggis-edit-categories})
\item Additional Details on Edit Instruction Distribution (Sec~\ref{app:frequent-words})
\item Additional Details on Human Evaluation (Sec~\ref{app:human-evaluation})
\item Additional Details on Automatic Evaluation (Sec~\ref{app:auto-eval})
\end{compactitem}

\subsection{Limitations and Future Work}
\label{app:limitations}
First, \dataset covers single-image and single-turn edits only. An interesting future research direction is to curate multi-turn editing dataset on top of \dataset images. Second, \dataset does not cover non-natural images such as info-graphics, thus limiting the scope of the model evaluation to natural images only. Third, although the LLM generated edit instructions (GGIS) have been curated thoughtfully to cover a diverse range of difficult and unique edit types per image-category, they might inherit LLM's inherent biases. However, despite these limitations, we find that our dataset reveals interesting gaps in model capabilities and directions for future research on model development. Thus, we believe it is a useful contribution for the research community

\subsection{Broader Impact}
\label{app:broader_impact}
Our work establishes a \emph{transparent} benchmark for image generation models by introducing a standardized, real-world dataset entirely independent of internet-scraped training corpora. Furthermore, we intentionally crafted non-trivial, creative, and complex edit instructions to stress-test current state-of-the-art models and provide a high-quality benchmark for future systems. Additionally, our Gemini-based autorater closely mirrors human judgment, offering a scalable evaluation mechanism that bypasses the steep financial and logistical bottlenecks of human judgement collection. Our work does not have any direct negative implications, but our work supports development of stronger image editing models, which in turn, could potentially be used for harmful applications.

\subsection{Computational Requirements}
\label{app:resources}
We utilized Gemini 3 Pro to generate GGIS edit instructions and Gemini 3 Flash to develop and evaluate the models' generations using the proposed TECCI autorater. All tasks relied on standard API calls, eliminating the need for dedicated hardware. The comprehensive automatic evaluation procedure, assessing 7,550 generations across five models with five runs each, required a total of 188,750 API calls. Additional API calls were consumed during the development of the autorater and the generation of GGIS edit instructions.

\subsection{Additional Details on Image Collection and Curation}
\label{app:image-curation}

The raw images themselves were taken, and thus owned, by a single individual (one of the authors), who has donated them under the CC-BY license, ensuring they are a stable evaluation set with easy access for the research community. Many existing image datasets are collections of links to images on the internet, and the targets of such links naturally disappear over time, reducing the value of the data considerably. However, anyone can download all the TECCI images and store them without worrying about future removals. This ensures that future comparisons will remain valid with respect to past ones as their is no degradation of the data over time. We hope this encourages others to use these as raw material for further challenge sets, such as multi image combination and editing tasks, reference to video, and so on. 

A few more notes on the collection and curation:

\begin{itemize}
    \item The images were taken generally to capture a fairly clean scene that often focus on one or two key subjects and avoids mess.
    \item This was made even cleaner by manual cropping of the images to remove extraneous material. This also included some rotation and warping, e.g. to ensure a painting was squared up post hoc to fix keystoning (trapezoidal warping).
    \item Any license plates were scrubbed manually.
    \item No close-ups of people were taken. In the event people were present in an image, they were blurred out if there was any risk of their face showing large enough. Faces were also scrubbed from any reflections (or the image was rejected if that was not possible).
\end{itemize}

\subsection{Full List of TECCI-GGIS Edit Types}
\label{app:ggis-edit-categories}
Table \ref{tab:ggis_edits} presents the full list of the TECCI-GGIS edit types, consisting of 5 edit types per image category, as discussed in Section \ref{sec: edit_instruction_curation}.

\begin{table*}[htbp]
\centering
\footnotesize
\caption{Edit types for each image category in TECCI-GGIS.
}
\label{tab:ggis_edits}
\begin{tabularx}{\textwidth}{|l|l|X|}
\toprule
\textbf{Image Cat.} & \textbf{Edit Type} & \textbf{Description of Edit Type} \\ \midrule
\multirow{5}{*}{Text}         & Text Content     & Changing the text to a different text in the same language. \\
                              & Translation      & Translating the text into a different language.                \\
                              & Text Style       & Changing the style of the text (e.g., font, theme) while preserving the text content and the background.  \\
                              & Background Style & Changing the style of the background (e.g., theme, new background) while preserving the text content and text style. \\
                              & Creative         & Creative and humorous changes to the text. \\

\hline
\multirow{5}{*}{Clock}        & Time           &  Changing the time displayed on the clock.         \\
                              & Text           &  Adding or removing text from the clock.           \\
                              & Appearance     &  Changing the appearance and material of the clock. \\  
                              & Image Style    &  Changing the style of the image (e.g., converting to comic style, converting to photorealistic image).                                               \\
                              & Creative       &   Creative and humorous changes to the clock.  \\         

\hline
\multirow{5}{*}{Vehicle}      & State and Motion &  Changing the state and motion of the vehicle (e.g., making an airplane fly if it is standstill). \\
                              & Text             &  Changing the text on the vehicle (e.g., brand name, graphic illustrations).  \\
                              & Appearance       &  Changing the appearance of the vehicle (e.g., color, texture, model).         \\        
                              & Environment      &  Changing the environment the vehicle is situated in (e.g., making a car drive on a mountain).                                              \\ 
                              & Creative         &  Creative and humorous changes to the vehicle.                                               \\ 

\hline
\multirow{5}{*}{Architecture} & Structural      & Structural changes to the architecture (e.g., adding or removing floors, wings). \\
                              & Viewpoint       & Showing a different view of the architecture (e.g., back view, aerial view, side view). \\
                              & Appearance      & Changing the appearance of the architecture (e.g., color, texture, adding intricate details, show the architecture in futuristic style). \\
                              & Environment     & Changing the environment the architecture is situated in (e.g., placing it in water instead of land, changing the scene from daytime to nighttime). \\
                              & Creative        &  Creative and humorous changes to the architecture shown in the picture. \\

\hline
\multirow{5}{*}{Art}          &   Semantic      & Semantic changes depending on the content of the art -- A) People: changing their pose, action, facial expression etc., B) Landscape: changing the count of trees, replacing a hut with a modern house etc., C) Close-up of hands: changing the pose of the hands, clasping of fingers etc. \\
                              &   Restoration   & Restoring and enhancing the art to improve the quality of the original illustration and make it clearer. \\
                              &   Color         &  Adding color to monochrome images. \\
                              &   Image Style   &  Changing the style of the image (e.g., converting to comic style, converting to photorealistic image). \\
                              &   Creative      &  Creative and humorous changes to the main character in the art. \\

\hline
\multirow{5}{*}{Animal}       &   Pose and Action  & Pose and action changes (e.g., making the dog lick the ball, making the cat stand-up on all fours, making the bird fly). \\
                              &   Position         &  Changing the position of the animal (e.g., moving one animal closer or farther away from the other, moving them in-front-of or behind another object).   \\
                              &   Animal Style     & Changing the style of the animal (e.g., changing their color, texture). \\
                              &   Image Style      & Changing the style of the image (e.g., converting to comic style, converting to photorealistic image). \\
                              &   Creative         & Creative and humorous changes to the animal. \\

\hline
\multirow{5}{*}{Nature}       &   Semantic    &  Changes to the content of the scene (e.g., adding or remove huts, trees, animals). \\
                              &   Scale       &   Scale changes to the image (e.g., zooming into a specific mountain, a specific bird, or zooming out to show a wider view of the scene). \\
                              &   Image Style &  Changes to the style of the image (e.g., converting it to a specific kind of painting). \\
                              &   Theme       &  Changing the theme of the image (e.g., changing the weather from sunny to stormy, changing the light to simulate sunrise or sunset casting long shadows).  \\
                              &   Creative    &  Creative and humorous changes to the nature elements shown in the picture.   \\

\bottomrule
\end{tabularx}
\end{table*}

\subsection{Additional Details on Edit Instruction Distribution}
\label{app:frequent-words}

In this section, we present a table of the most frequent words (Table~\ref{tab:top_100_words}) for both the TECCI-GGIS and TECCI-IRCS subsets. This complements the lexical diversity analysis discussed in Section~\ref{sec: dataset_analysis}. 

\begin{table*}[h]
\centering
\caption{Top 100 most frequent words in TECCI-GGIS and TECCI-IRCS (excluding stop words).}
\label{tab:top_100_words}
\small
\begin{tabular}{rll|rll}
\toprule
\textbf{Rank} & \textbf{GGIS} & \textbf{IRCS} & \textbf{Rank} & \textbf{GGIS} & \textbf{IRCS} \\
\midrule
1 & change (2757) & change (124) & 51 & gold (305) & people (23) \\
2 & background (1541) & make (116) & 52 & looking (301) & holding (23) \\
3 & replace (1508) & right (111) & 53 & view (300) & letter (22) \\
4 & image (1261) & left (106) & 54 & building (300) & sitting (22) \\
5 & text (1246) & add (93) & 55 & modify (297) & instead (21) \\
6 & white (1148) & hand (85) & 56 & translate (296) & expression (21) \\
7 & add (1036) & painting (79) & 57 & purple (293) & quality (21) \\
8 & style (855) & show (77) & 58 & metal (272) & new (21) \\
9 & black (847) & put (73) & 59 & hand (271) & photograph (21) \\
10 & keep (812) & turn (69) & 60 & left (269) & around (21) \\
11 & green (772) & remove (62) & 61 & translucent (269) & drawing (20) \\
12 & bright (662) & top (60) & 62 & else (267) & red (20) \\
13 & blue (651) & real (59) & 63 & pink (267) & wall (20) \\
14 & large (643) & man (53) & 64 & colorful (265) & write (20) \\
15 & dark (636) & like (53) & 65 & show (255) & create (20) \\
16 & small (631) & white (49) & 66 & exactly (254) & woman (20) \\
17 & red (608) & image (49) & 67 & remove (252) & keep (19) \\
18 & texture (579) & scene (47) & 68 & anything (249) & old (19) \\
19 & stone (579) & word (46) & 69 & face (241) & using (19) \\
20 & color (567) & two (45) & 70 & right (240) & building (19) \\
21 & entire (525) & one (44) & 71 & directly (237) & use (19) \\
22 & letters (512) & bottom (43) & 72 & deep (228) & time (19) \\
23 & like (492) & black (39) & 73 & realistic (227) & finger (19) \\
24 & glass (485) & written (36) & 74 & colored (225) & statue (19) \\
25 & thick (481) & looking (36) & 75 & scene (223) & showing (18) \\
26 & top (477) & swap (36) & 76 & letter (217) & middle (18) \\
27 & giant (475) & front (34) & 77 & high (215) & figure (18) \\
28 & glowing (457) & car (33) & 78 & outlines (215) & three (18) \\
29 & convert (445) & words (33) & 79 & water (214) & camera (17) \\
30 & visible (436) & standing (33) & 80 & solid (212) & clock (17) \\
31 & look (434) & side (33) & 81 & bottom (205) & page (17) \\
32 & car (431) & style (32) & 82 & sign (202) & original (17) \\
33 & appear (413) & blue (31) & 83 & paint (202) & behind (17) \\
34 & transform (403) & also (31) & 84 & grey (201) & panel (17) \\
35 & make (401) & sticky (30) & 85 & place (201) & shown (16) \\
36 & surface (400) & look (30) & 86 & light (200) & massive (16) \\
37 & made (395) & number (30) & 87 & door (200) & keeping (15) \\
38 & clock (394) & book (30) & 88 & wheels (200) & speech (15) \\
39 & sky (372) & cat (29) & 89 & bold (199) & based (15) \\
40 & yellow (370) & letters (29) & 90 & window (195) & person (15) \\
41 & polished (368) & background (29) & 91 & weathered (193) & large (15) \\
42 & font (357) & note (28) & 92 & metallic (188) & rotate (15) \\
43 & vibrant (344) & color (28) & 93 & aerial (188) & sure (15) \\
44 & side (338) & yellow (27) & 94 & silver (184) & design (15) \\
45 & neon (331) & colors (25) & 95 & emerald (183) & cover (15) \\
46 & word (330) & move (25) & 96 & detailed (183) & comic (15) \\
47 & orange (325) & head (25) & 97 & center (182) & square (15) \\
48 & wooden (322) & face (24) & 98 & smooth (180) & bubble (14) \\
49 & two (308) & photo (23) & 99 & pair (175) & life (14) \\
50 & wall (307) & high (23) & 100 & carved (171) & lines (14) \\
\bottomrule
\end{tabular}
\end{table*}

\clearpage
\subsection{Additional Details on Human Evaluation}
\label{app:human-evaluation}

\begin{figure*}[h]
    \centering
    \includegraphics[width=0.95\textwidth]{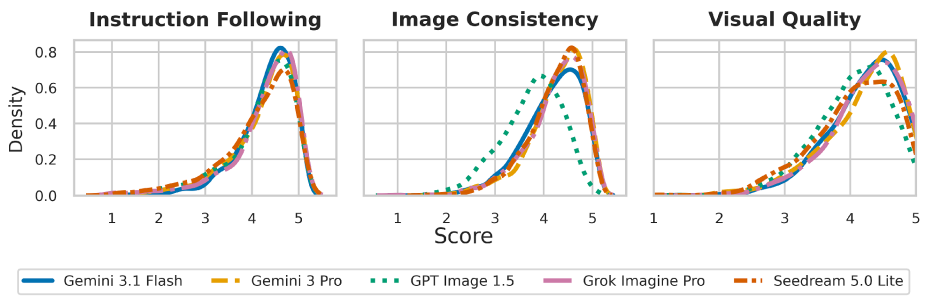}
    \caption{Distribution of human evaluation scores across the three evaluation criteria.}
    \label{fig:he_kde_scores}
\end{figure*}

The following section provides a detailed statistical overview of the single-side human evaluation study. 
Figure \ref{fig:he_kde_scores} illustrates the distribution of mean raw human judgement scores (mean across the five raters) obtained utilizing a Kernel Density Estimation (KDE) fit. Given that the evaluation involves state-of-the-art generative models, the mean raw scores consistently hover around 4 points across all assessed criteria. However, to better capture the nuances of high-fidelity performance and ensure the generation of accurate, complete edits, we prioritize reporting thresholded accuracy alongside the comprehensive \overall metric within the main body of the paper.

\begin{figure}[ht!]
     \centering
     \includegraphics[width=1\textwidth]{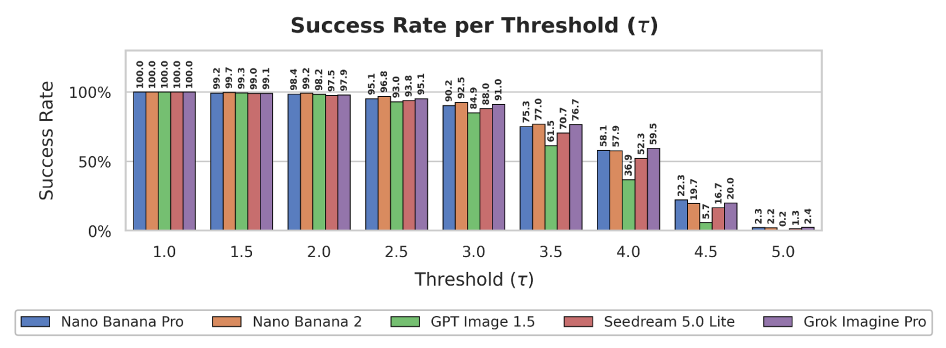}
     \caption{Distribusion of \overall scores for different thresholds $\tau$}
     \label{fig:sr_per_threshold}
\end{figure}

Figure \ref{fig:sr_per_threshold} illustrates the per-model \overall score for different choices of thresholds $\tau$. As shown, selecting a threshold below 3.0 yields trivial results, as the majority of mean scores exceed this value. Conversely, a threshold of 5.0 is overly strict, resulting in compressed variance and low success rates across the board. Generally, model performance trends remain consistent for thresholds within the $[3, 4.5]$ range, with only modest variations in ranking. We ultimately selected a threshold of $\tau=4.5$ after a manual inspection of samples from all models, confirming that it balances a high standard for quality with sufficient variance to differentiate between the tested models.

\clearpage
\subsubsection{Single Side Annotation Template}
\label{app:human-eval-sinlge-temp}

\begin{figure}[h!]
    \centering
    \includegraphics[width=0.8\textwidth]{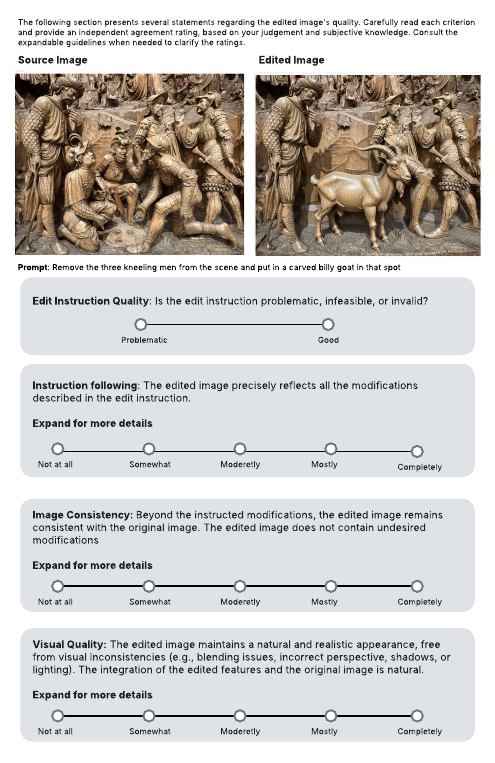}
    \caption{Annotation interface for the single-side evaluation task.}
    \label{fig:single_side_annotation_interface}
\end{figure}

Figure \ref{fig:single_side_annotation_interface} presents an illustration of the annotation interface used by human raters for the single-side evaluation setup. As mentioned in the main paper, annotators underwent an extensive training procedure that is not included in this report. First, annotators verified that the edit instructions were valid, i.e., relevant to the source image, feasible, and appropriate. They then ranked the generated images along the three individual evaluation criteria according to our detailed guidelines. Each sample was evaluated by five independent annotators.

For the edit instruction verification, 98.1\% of the instructions (1289 out of 1315) were marked as valid by all five annotators. For the remaining 1.9\% of the instructions (26 out of 1315), only one annotator marked the instruction as invalid resulting in 80\% agreement across the annotators.
This high level of consensus demonstrates that the edit instructions are of high quality.

\clearpage
\subsubsection{Side-by-Side Annotation Template}
\label{app:human-eval-sxs-temp}

\begin{figure}[h!]
    \centering
    \includegraphics[width=0.95\textwidth]{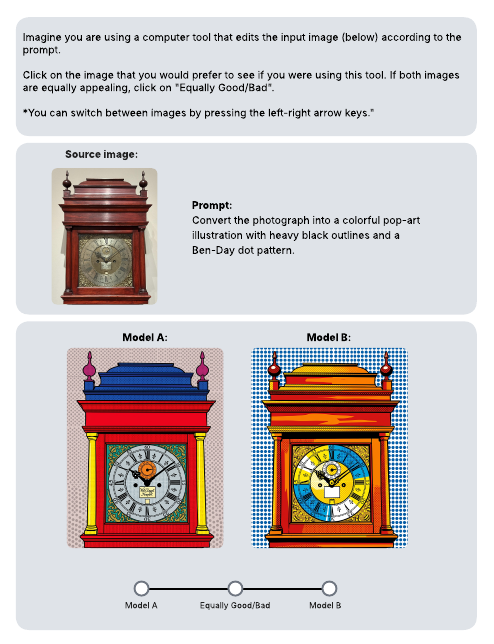}
    \caption{Annotation interface for the side-by-side evaluation task. Each task presents the source images, edit instruction and two edited images from different models. Raters are asked to provide an overall preference score.}
    \label{fig:sxs_annotation_interface}
\end{figure}

Figure \ref{fig:sxs_annotation_interface} presents the side-by-side annotation interface used by human raters to score pairs of edited images sampled from the tested models. As elaborated in the main paper, the side-by-side evaluation focuses on an ``overall quality'' impression, as capturing multiple criteria across all pairs is out of scope. This evaluation complements the single-side evaluation and provides a strong overall preference signal.

\subsection{Additional Details on Automatic Evaluation}
\label{app:auto-eval}

\begin{figure}[h!]
    \centering
    \includegraphics[width=0.99\textwidth]{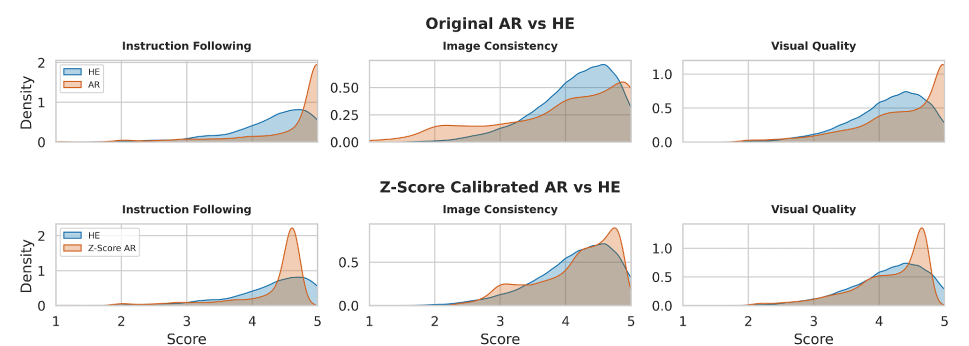}
    \caption{Score distributions for Autorater (AR) and Human Evaluation (HE) before (top) and after (bottom) autorater score normalization..}
    \label{fig:ar_he_dists}
\end{figure}

This section presents supplementary results regarding the TECCI image editing autorater. As discussed in the main paper, we observed significant discrepancies between the score distributions of the autorater and the human-based evaluation. To mitigate these distributional discrepancies, we applied z-score matching to align the two score distributions, normalizing them independently per criterion \citep{li2026grading}. The autorater scores were transformed according to: $s_{AR}^{'} = \mu_{HE} + \left( \frac{s_{AR} - \mu_{AR}}{\sigma_{AR}} \right) \cdot \sigma_{HE}$. The normalization statistics were calculated using the human-annotated subset and applied to all \dataset autoratings. Figure \ref{fig:ar_he_dists} presents the original scores distribution (top row) and the normalized scores distribution (bottom row).

Table \ref{tab:ar_corr} outlines the correlation between human evaluations and the autorater's scores, broken down by model and evaluation criteria. The data reveals a moderate correlation, which aligns with the inherently complex, multi-dimensional, and subjective nature of evaluating image editing. For benchmarking purposes, we also include the human-to-human correlation coefficients for each setup. Notably, the inter-human agreement is also modest—further highlighting the subjectivity of the task. Overall, the autorater-to-human correlation is highly competitive and does not fall significantly short of the human-to-human baseline.

 We utilized Chain-of-Thought prompting \citep{wei2022chain} when developing the autorater prompt. We observed that obliging the model to perform a granular, step-by-step analysis before assigning a numerical value yields superior evaluation quality compared to direct scoring. Figure \ref{fig:prompt} provides the detailed autorater prompt, we also provide inference script using this prompt.

\begin{table}[th]
\centering
\caption{Krippendorff's Alpha Coefficients: Human to Human vs. Autorater to Human Comparison}
\begin{tabular}{lcccccc}
\toprule
& \multicolumn{3}{c}{\textbf{Human to Human}} & \multicolumn{3}{c}{\textbf{Autorater to Human}} \\
\cmidrule(lr){2-4} \cmidrule(lr){5-7}
\textbf{Model} & \textbf{IF} & \textbf{IC} & \textbf{VQ} & \textbf{IF} & \textbf{IC} & \textbf{VQ} \\ 
\midrule
Nano Banana Pro   & 0.585 & 0.600 & 0.545 & 0.551 & 0.551 & 0.591 \\
\nbtwo     & 0.660 & 0.624 & 0.609 & 0.656 & 0.520 & 0.635 \\
Grok Imagine Pro  & 0.625 & 0.599 & 0.564 & 0.537 & 0.492 & 0.608 \\
\seedream & 0.573 & 0.633 & 0.506 & 0.494 & 0.579 & 0.574 \\
\gpt     & 0.617 & 0.504 & 0.558 & 0.508 & 0.481 & 0.627 \\ 
\midrule
\textbf{Mean}      & \textbf{0.612} & \textbf{0.592} & \textbf{0.556} & \textbf{0.549} & \textbf{0.524} & \textbf{0.607} \\
\bottomrule
\end{tabular}
\label{tab:ar_corr}
\end{table}

\begin{figure*}[t]
\begin{tcolorbox}[
    colback=gray!5,
    colframe=gray!50,
    title=Evaluation Prompt Template,
    arc=4pt,
    boxsep=2pt, 
    left=4pt, right=4pt, top=2pt, bottom=2pt]
\small 

\textbf{Task Description} \\
You are an expert visual inspector, and analyst of image editing quality. In this task, you will be asked to evaluate the quality of an edited image based on a few specific criteria. You will be provided with a source image, an edited image, and an edit instruction.

First, carefully examine the source image, the edited image, and the edit instruction. Then, provide a detailed analysis in the ``evaluation\_analysis'' field. Your analysis MUST include:
\begin{enumerate}
    \setlength{\itemsep}{0pt} 
    \setlength{\parskip}{0pt}
    \item \textbf{Requested changes:} List every distinct modification requested by the edit instruction.
    \item \textbf{Delivered changes:} For each requested modification, state whether it was correctly applied, partially applied, or not applied.
    \item \textbf{Unintended changes:} List any changes you observe in the edited image that were NOT requested by the instruction. Consider changes in background, lighting, color palette, object positions, facial features, textures, and any other visual element.
    \item \textbf{Overall impression:} A brief summary of the edit quality.
\end{enumerate}

Finally, based on your analysis, provide a rating and a short justification for each criterion. Follow those instructions carefully when evaluating the image editing quality:
\begin{enumerate}
    \setlength{\itemsep}{0pt} 
    \setlength{\parskip}{0pt}
    \item Each rating must be an integer between 1 and 5 (1 is not at all, 5 is completely).
    \item Each justification must be a short sentence.
    \item Judge each criterion separately, and do not let the rating of one criterion influence the another.
    \item As image editing quality is subjective, answer honestly, according to your best judgement and as an average human visual inspector would.
    \item \textbf{Be strict in your ratings.} Only give a rating of 5 if the criterion is perfectly met.
\end{enumerate}

\tcbline

\textbf{Evaluation Criteria}

\vspace{0.3em}
\textbf{1. Instruction Following:} Does the edited image precisely reflect all modifications described? \\
\hspace*{1.5em} \textit{Note: Your rating must be consistent with the ``Delivered changes'' checklist above. If any requested change is missing or only partially applied, the rating cannot be 5.} \\
\hspace*{1.5em} \textbf{Completely (5):} Accurately and completely adheres to the modifications. \\
\hspace*{1.5em} \textbf{Mostly (4):} Reflects most modifications with minor discrepancies. \\
\hspace*{1.5em} \textbf{Moderately (3):} Some modifications are met; some do not match. \\
\hspace*{1.5em} \textbf{Somewhat (2):} Some modifications reflect, but key elements are missing or inaccurate. \\
\hspace*{1.5em} \textbf{Not at all (1):} Identical to the original; the edit has not been applied.

\vspace{0.3em}
\textbf{2. Image Consistency:} Beyond the instructions, does the image remain consistent with the original? \\
\hspace*{1.5em} \textit{Note: Your rating must be consistent with the ``Unintended changes'' list above. If there are significant unintended changes, the rating cannot be 5.} \\
\hspace*{1.5em} \textbf{Completely (5):} Fully consistent; preserves all features not intended to be changed with high accuracy. \\
\hspace*{1.5em} \textbf{Mostly (4):} Largely retains consistency; minor unintended changes. \\
\hspace*{1.5em} \textbf{Moderately (3):} Some features preserved, others incorrectly modified. \\
\hspace*{1.5em} \textbf{Somewhat (2):} Essential features that were not meant to be modified have been altered. \\
\hspace*{1.5em} \textbf{Not at all (1):} Unmodified parts are completely different or unrecognizable.

\vspace{0.3em}
\textbf{3. Visual Quality:} Is the integration seamless, natural, and free from artifacts? \\
\hspace*{1.5em} \textbf{Completely (5):} Blends perfectly and feels completely natural. \\
\hspace*{1.5em} \textbf{Mostly (4):} Blends well; minor alterations visible when paying close attention. \\
\hspace*{1.5em} \textbf{Moderately (3):} Somewhat consistent, but it is noticeable that the image has been edited. \\
\hspace*{1.5em} \textbf{Somewhat (2):} Edit doesn't blend well; differences are significant. \\
\hspace*{1.5em} \textbf{Not at all (1):} Modified in a way that is completely unrecognizable.

\tcbline
\textbf{Output Format} \\
Please provide your evaluation in the following JSON format:
\vspace{-0.5em} 
\begin{verbatim}
{
  "evaluation_analysis": "...",
  "instruction_following": {
        "rating": number in [1, 5],
        "reason": "Short justification for this rating."
    },
  "image_consistency": {
        "rating": number in [1, 5],
        "reason": "Short justification for this rating."
    },
  "visual_quality": {
        "rating": number in [1, 5],
        "reason": "Short justification for this rating."
    }
}
\end{verbatim}
\end{tcolorbox}
\caption{TECCI image editing autorater prompt.}
\label{fig:prompt}
\end{figure*}

\clearpage
\subsection{Additional Examples}
\label{app:additional_samples}

\begin{figure}[h!]
    \centering
    \includegraphics[trim=0 4cm 0 0, clip, width=0.99\textwidth]{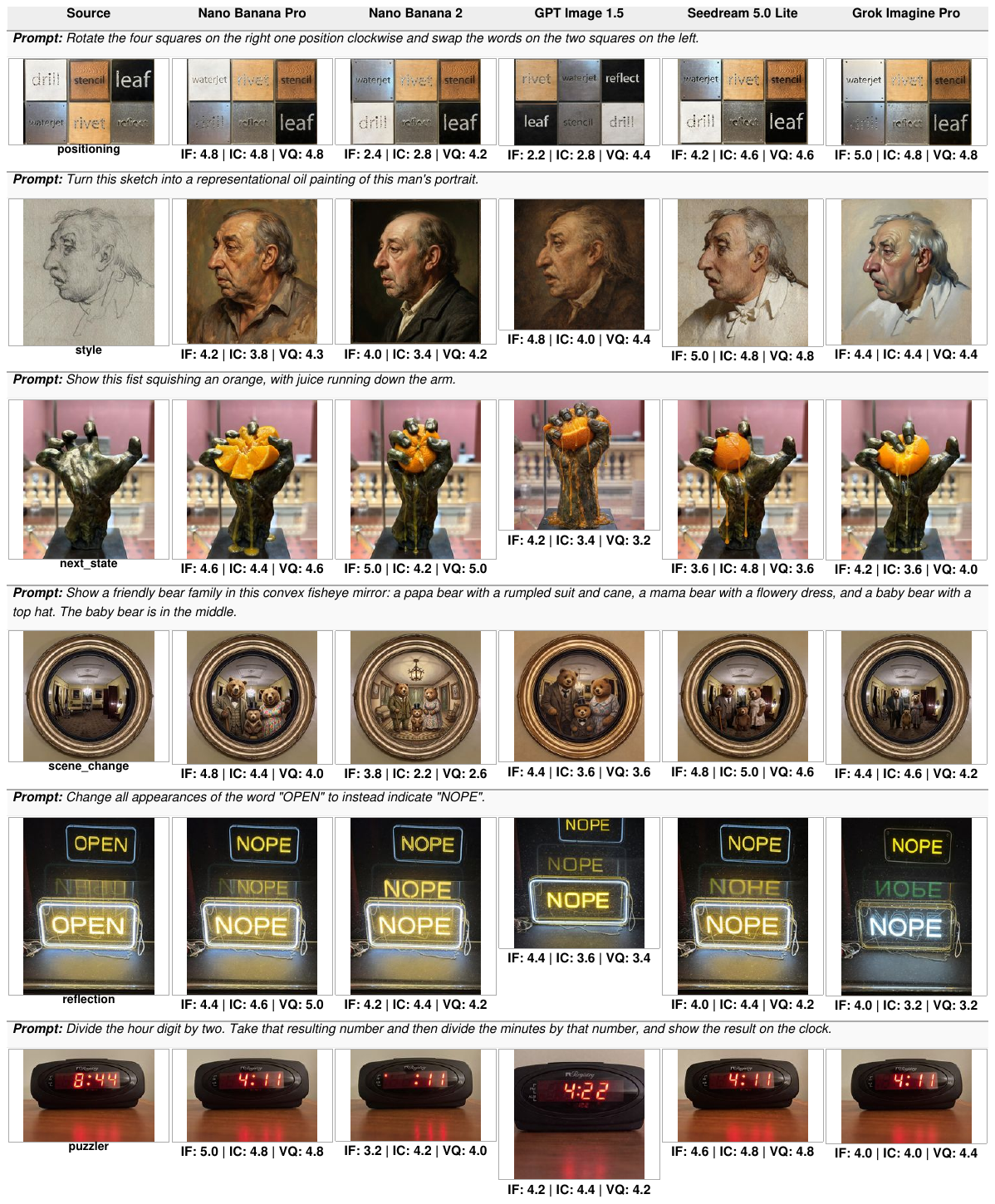}
    \caption{TECCI-IRCS examples with mean human-based scores.}
    \label{fig:ircs_examples}
\end{figure}

\begin{figure}[h!]
    \centering
    \includegraphics[trim=0 4cm 0 0, clip, width=0.99\textwidth]{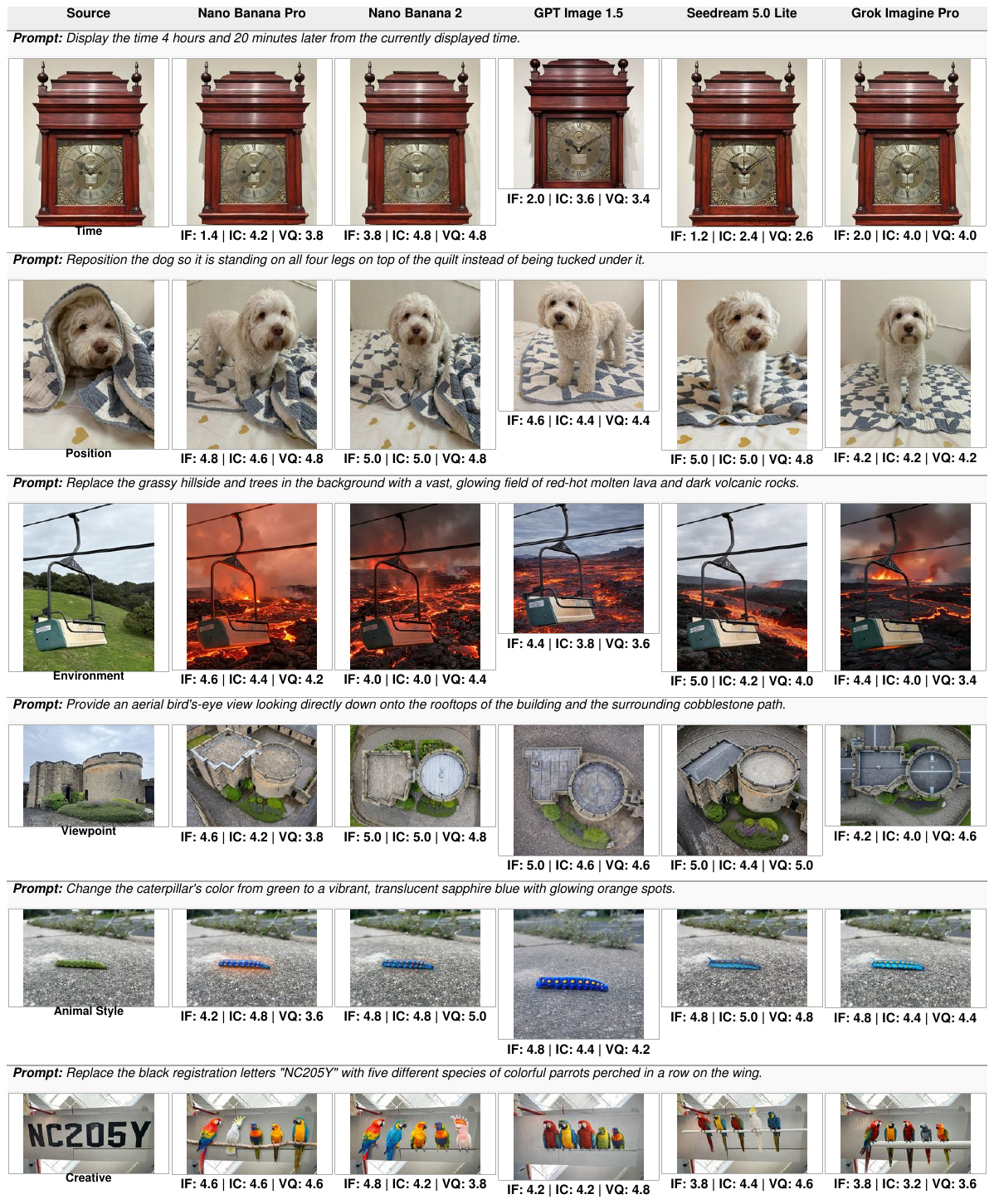}
    \caption{TECCI-GGIS examples with mean human-based scores.}
    \label{fig:ircs_examples}
\end{figure}